IJACT-16-2-1

# Comparison of various image fusion methods for impervious surface classification from VNREDSat-1

Hung V. Luu [†*], Manh V. Pham[†], Chuc D. Man[†], Hung Q. Bui[†], Thanh T.N. Nguyen[†]

[†] *Center of Multidisciplinary Integrated Technologies for Field Monitoring*
*E-mail: hunglv@fimo.edu.vn*

### Abstract

*Impervious surfaces are important indicators for urban development monitoring. Accurate mapping of urban impervious surfaces with observational satellites, such as VNREDSat-1, remains challenging due to the spectral diversity not captured by an individual PAN image. In this article, five multi-resolution image fusion techniques were compared for the task of classifting urban impervious surfaces. The result shows that for VNREDSat-1 dataset, UNB and Wavelet tranformation methods are the best techniques in reserving spatial and spectral information of original MS image, respectively. However, the UNB technique gives the best results when it comes to impervious surface classification, especially in the case of shadow areas included in non-impervious surface group.*

## 1. INTRODUTION

Impervious surface are mainly artificial structure that are covered by impenetrable material and has been recognized as an important indicator in urban development monitoring [1][2]. The increasing availability of Very High Resolution (VHR) imagery provides a great opportunity for detail impervious surface mapping in urban area.

Due to cost and complexity issues, recently launched VHR satellites often provide us a PANchromatic (PAN) images with finer spatial resolution than MultiSpectral (MS) images. However, MS images have higher spectral resolution than PAN images, thus were more applicable for pixel based classification task. Since combining the output from different sensors makes the best use of data obtained from existing satellites [3], the good fusion of the MS and PAN images is able to utilize the advantages of both. A fusion image are preserving the spectral resolution of MS images spatial resolution of PAN image, which can avoid mixed pixel problem in classification of sparse resolution MS images.

The goal of the article is to investigate a fusion method that would increase spatial resolution without degrading spectral discrimination for mapping of impervious surface in urban area. Five widely used multi-resolution merging methods are compared. In our experiments, we used VNREDSat-1 images over Saigon Port area in Ho Chi Minh city, Vietnam. The VNREDSat-1 was launched in May 2013 as the first satellite of Vietnam aiming to capture high resolution image for natural resources, environment and disaster monitoring and management.

In the next section, images fusion is discussed. Results of impervious surface classification using different fusion images are given in Section 3. Finally, conclusion and future work is presented.

## 2. MULTI RESOLUTION IMAGE FUSION

The study area is Saigon Port area at Ho Chi Minh city, Vietnam. It is one of the most urbanized areas in the city characterized by various impervious artificial construction types including small roads, large roads, bridges, rooftops, etc … Besides, non-impervious surface including trees, park, water, … are also presented and thus creating the diversity of area. The dataset is a VNREDSat-1 image recorded in Jan 30th, 2014 with detailed information presented in Table 1. VNREDSat-1 provides a dataset with a PAN band at 2.5 m of spatial resolution and MS images of four bands including Red, Green, Blue, and Near InFrared (NIR) at 10 m spatial resolution.

**Table 1. The spectrum of the VNREDSat-1 bands**

| Band Name | Name | Spatial Resolution (m) | Central wavelength (µm) | Wavelength (µm) |
|-----------|------|------------------------|-------------------------|-----------------|
| Multispectral | Blue | 10 | 490 | 0.45 – 0.52 |
| | Green | 10 | 550 | 0.53 – 0.59 |
| | Red | 10 | 660 | 0.625 – 0.695 |
| | Nir | 10 | 830 | 0.76 – 0.89 |
| Panchromatic | - | 2.5 | 600 | 0.45 – 0.75 |

Image fusion can be done at three levels including pixel, feature and decision level. In this study, we consider fusion at the pixel level with five widely used algorithms including IHS [4], PCA [5, 6], Gram-Schmidt (GS) [7], Wavelet transform [8, 9] and University of New Brunswick (UNB) method [10].

For each fused image, true and false color composites were produced and visually inspected. The visual analysis includes the following controls: existence of color distortion locally or globally in the image, existence of color tonality differences, detection of linear distortion in roads, buildings, bridges, soil .. and general appearance of the image (brightness, contrast, etc ...). Visual comparison (qualitative metrics) of different five fused images shows that UNB is the best one which reserves the representation of object details, then IHS and GS fusion images. Resulted images by PCA and Wavelet transformation make blurred details (Figure 1).

For qualitative assessment, a series of quality metrics was used to evaluate the spatial and spectral fidelity between fused images and original MS data. We consider Bias, Difference In Variance (DIV), Correlation Coefficient (CC), ERGAS, the Universal Image Quality Index (UIQI), Relative Average Spectral Error (RASE), Root Mean Square Error (RMSE) and Entropy following definitions in[11-15]. Those metrics were calculated in each of four bands and then averaged (Table 2). Wavelet transform is the best technique keeping spectral characteristics of the original MS image (i.e. DIV, CC, ERGA, UIQI, RMSE are smallest) and then, IHS, PCA, UNB and GS are following.

**Table 2 Quantitative assessment of VNREDSat-1 fusion images**

| | **Bias**<br>*(Ideal values: 0)* | **DIV**<br>*(Ideal values: 0)* | **CC**<br>*(Ideal values: 1)* | **ERGA**<br>*(Ideal values: 0)* | **RASE**<br>*(Ideal values: 0)* | **UIQI**<br>*(Ideal values: 1)* | **RMSE**<br>*(Ideal values: 0)* |
|---|---|---|---|---|---|---|---|
| **PCA** | 0.006 | -0.193 | 0.789 | 7.667 | 0.496 | 0.774 | 0.028 |
| **GS** | -0.015 | -0.141 | 0.881 | 6.627 | 0.267 | 0.863 | 0.021 |
| **IHS** | -0.014 | 0.058 | 0.911 | 5.056 | 0.198 | 0.910 | 0.016 |
| **Wavelet** | -0.008 | 0.049 | 0.958 | 3.576 | 0.233 | 0.957 | 0.013 |
| **UNB** | -0.014 | 0.085 | 0.896 | 5.487 | 0.230 | 0.894 | 0.018 |

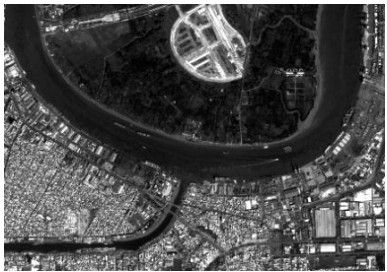

a- Original Pan

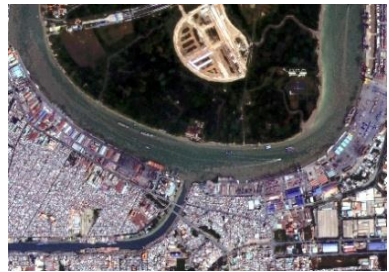

b- IHS

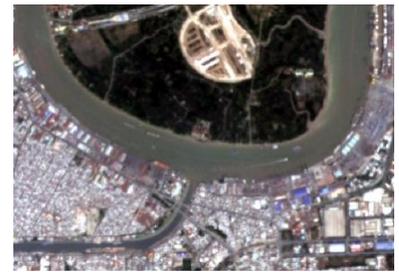

c- PCA

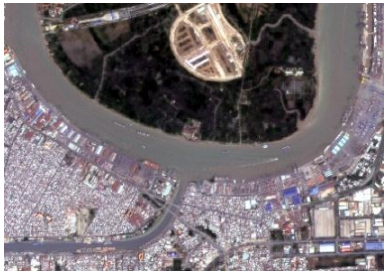

d- GS

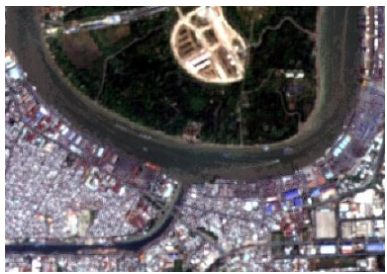

e- WL

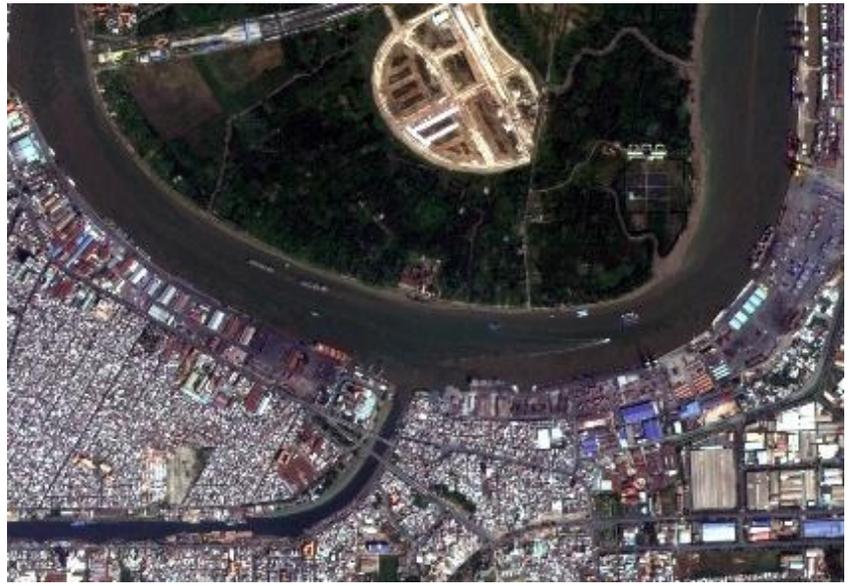

f- UNB

**Figure 1. VNREDSat-1 2.5m original PAN and generated fusion images**
**a) PAN b) HIS c) PCA d) GS e) WL f) UNB**

## 3. IMPERVIOUS CLASSIFICATION

In this section, we investigate impevious surface classification from different fusion images in which UNB, IHS and GS can keep spatial consistency while Wavelet, IHS and PCA reserve spectral characteristics for the VNREDSAT 1 image.

Impervious surface (IS) and non-impervious surface (NIS) are combination of various land cover types. Impervious surface can be made up of dark impervious surface (DIS) and bright impervious surface (BIS). Non-impervious surface consists of diverse materials including vegetation (VEG), water body (WAT), bare soil/sand (BSS), and shaded area (SHA).

In this study, a two-step approach was employed. Firstly, different land cover classes were grouped into impervious and non-impervious surfaces for classification. PAN and 5 fusion images are used to analyze the effectiveness of different fusion methods for classification task. Secondly, the best fusion dataset will be further analyzed for detail land cover.

Ground truth data for each land cover class in Saigon Port area have been selected manually from the VNREDSat-1 image. For each class, we selected 500 samples in which a half   (i.e. 250 samples) is for training and the rest is for testing. For the classification, the Support Vector Machine was used with the Gaussian Radial Basis function for the kernel and the training parameters estimated by a grid-search on each dataset.

Table 3 shows the overall accuracy of impervious and non-impervious classification using different fusion images. Most fusion images improve the accuracy in compare to original PAN data (84.4%) with exception of Wavelet (77.9%). The test accuracy improved significantly when using UNB image (89.6%) follow by PCA (89.1%) and GS (87.1%). In fact, different fusion images have smaller difference in spectral presentation (Table 2) than spatial presentation (Figure 1). The classification result seems strongly related to spatial than spectral aspects presented in fusion images and impevious and non-impervious classes themselves.

**Table 3. Confusion matrix and overall accuracy impervious/non-impervious classification**

|  | PAN | | PCA | | GS | | IHS | | Wavelet | | UNB | |
|---|---|---|---|---|---|---|---|---|---|---|---|---|
|  | IS | NIS | IS | NIS | IS | NIS | IS | NIS | IS | NIS | IS | NIS |
| **IS** (# pixels) | 420 | 80 | 451 | 49 | 417 | 83 | 398 | 102 | 306 | 194 | 464 | 36 |
| **NIS** (# pixels) | 172 | 828 | 128 | 872 | 114 | 886 | 136 | 864 | 130 | 870 | 139 | 861 |
| **Accuracy** (%) | 84.4 | | 89.1 | | 87.1 | | 84.5 | | 77.9 | | 89.6 | |

The further investigation on impevious and non-impervious classes is carried out on UNB fusion image and PAN image. The result is listed in

Table **4**. Several important findings can be observed. First, several impervious and non-impervious classes are easily confused when using only PAN image. For instance, 66 pixels of BIS are classified as BSS, 100 pixels of VEG are classified as DIS, 43 pixels of BSS are classified as BIS and 29 pixels of SHA are classified as DIS. Moreover, classes in some group (impervious or non-impervious) are also confused. 34 pixels of DIS are classified as BIS, 103 and 114 pixels of SHA are classified as VEG and WAT respectively. However, after combining PAN with MS using the UNB method, the mistakes are dramatically reduced since all pixels of DIS, VEG, WAT, BSS are classified correctly. The ineffective performance of UNB just happen by classification of SHA the class.

**Table 4. Confusion matrix for detail classification using UNB and PAN images**

| | UNB | | | | | | PAN | | | | | |
|---|---|---|---|---|---|---|---|---|---|---|---|---|
| | BIS | DIS | VEG | WAT | BSS | SHA | BIS | DIS | VEG | WAT | BSS | SHA |
| **BIS** | 240 | 0 | 0 | 0 | 10 | 0 | 184 | 0 | 0 | 0 | 66 | 0 |
| **DIS** | 0 | 250 | 0 | 0 | 0 | 0 | 34 | 202 | 14 | 0 | 0 | 0 |
| **VEG** | 0 | 0 | 250 | 0 | 0 | 0 | 0 | 100 | 134 | 16 | 0 | 0 |
| **WAT** | 0 | 0 | 0 | 250 | 0 | 0 | 0 | 0 | 2 | 248 | 0 | 0 |
| **BSS** | 0 | 0 | 0 | 0 | 250 | 0 | 43 | 0 | 0 | 0 | 207 | 0 |
| **SHA** | 15 | 116 | 0 | 22 | 0 | 97 | 0 | 29 | 103 | 114 | 0 | 4 |

## 4. CONCLUSION

In the study, five different image fusion methods, including PCA, GS, IHS, Wavelet and UNB, were firsly investigated to produce higher resolution MS images, then used for impervious surface classification. The fusion results show that for VNREDSat-1 dataset, UNB and Wavelet tranform methods are the best techniques reserving spatial and spectral information of original MS image, respectively. The application of fusion images for current impervious classification points out strong relationship of classification to spatial than spectral aspects presented in fusion images. Therefore, the UNB is the best candidate for impervious surface classification, especially in the case of shadow area included in non-impervious surface group.

## ACKNOWLDEGEMENT

This work was supported by Vietnamese Science and Space Program   in 2014-2015.

## REFERENCESE